# A Single-Pass Classifier for Categorical Data

Kieran Greer, Distributed Computing Systems, Belfast, UK.
http://distributedcomputingsystems.co.uk
Version 1.3

**Abstract –** This paper describes a new method for classifying a dataset that partitions elements into their categories. It has relations with neural networks but a slightly different structure, requiring only a single pass through the classifier to generate the weight sets. A grid-like structure is required as part of a novel idea of converting a 1-D row of real values into a 2-D structure of value bands. Each cell in any band then stores a distinct set of weights, to represent its own importance and its relation to each output category. During classification, all of the output weight lists can be retrieved and summed to produce a probability for what the correct output category is. The bands possibly work like hidden layers of neurons, but they are variable specific, making the process orthogonal. The construction process can be a single update process without iterations, making it potentially much faster. It can also be compared with k-NN and may be practical for partial or competitive updating.

## 1   Introduction

This paper describes a new method for classifying a dataset that partitions elements into different categories. It has relations with neural networks but a slightly different structure, requiring only a single pass through the classifier to generate the weight sets. It also uses self-organisation and there is the possibility of competitive learning. A grid-like structure is required for storing the information, along with a novel idea of converting a one-dimensional row of real values into a two-dimensional structure of value bands. Each data value is placed in a grid row, where each grid row is a graded band, representing an input variable. Each cell in the band can store an individual set of weights to represent its own importance and its relation to each output category. There is therefore an additional band layer between the input variable and the output categories, rather like a neural network hidden layer. Different to a neural network however is the fact that the band layers are variable (column) specific and not data (row) specific, making the variables orthogonal. For any input that needs to be categorised, all of the output weight value sets for each cell that the input falls into, can be retrieved and summed, to produce a probability for what the correct output category is. So the relative importance of each input data point to the output is distributed to each cell. The construction process does not require iterative updates, making it potentially much faster.

The rest of this paper is organised as follows: section 2 describes some related work and section 3 describes the new classifier architecture. Section 4 gives an example scenario for how it might be used. Section 5 gives the results of some initial tests. Section 6 is a discussion suggesting future work, while section 7 gives some conclusions to the work.

## 2   Related Work

The influence for the new classifier has come from different areas. The classifier is not obviously a neural network because of its fixed grid-like structure. It was however inspired in a small way by the topology feature of the Self-Organising Map [5][16] and looking a bit deeper reveals that it contains the essential features of a typical neural network. It is possibly more visual than what is normal, but the SOM topology can also present the input data as a 2-D array and has been used to learn visual elements. There is a new structure however and it also requires a different training algorithm. Another example might be [9] (or the related Neocognitron) again because of the visual functionality. This new model is not obviously competitive like the SOM and it is also supervised, requiring both the input and desired output values during training. The training process itself is automatic and requires relatively few stages. However, as suggested later, there might be a possibility for a second training stage that would be more competitive. For online learning, this would be important. The new classifier also uses a direct association approach, rather like an associative network [21], but it is probably not the same as that either. It still tries to generalise over the input by deconstructing the datasets and not to produce an exact memory-recall mapping.

For comparisons with earlier work, a grid-like structure was also used in [11] to try to represent bits of a problem or solution, but using a different type of system and so it is not too similar. Critical was the idea of comparing wave shapes from [10], where in that case, the shape itself would help to determine what input sets to classify together, through a subsequent combining of related synapses. This is also a different type of project, but was a source of some of the idea and is mentioned for background information. As the author is interested in brain-like/cognitive models, there was a small amount of influence from the biological/genetic gel classifiers. Even if they use molecule size, they still classify into bands. The result might also be important for a recent and controversial paper [4][22] which suggests that neurons inside the brain store a memory of what synapses they should form. Previously, this information was thought to be in the synapses themselves. However they state that: 'Yet there's no known mechanism by which a neuron could store a molecular 'map' of its own connections and their differing strengths', where a pattern of different strengths is required. Could the banded structure store that map, especially if a fixed chemical size determines it?

Single-pass classifiers have been developed previously and are particularly useful when memory or time is at a premium; for example, for very large or online dataset clustering [1]. In that case, some type of incremental update can be applied, with new datasets being incorporated using single passes through the new dataset values. Other examples might include [17][19]. While these classifiers can note similar reasons or uses, they look different in their exact construction processes. Some common problems still exist however, such as skewed data or the sampling size. Regularised Discriminant Analysis (RDA) [8] has been used previously and is designed to tackle that problem. For the tests of this paper, there was more of a manual and intuitive process to re-balancing the datasets. One dataset tested was the Wine [7] dataset. The UCI [23] web page states that the classes are separable, but only RDA has achieved 100% correct classification. Other classifiers achieved: RDA 100%, QDA 99.4%, LDA 98.9%, 1NN 96.1% (z-transformed data) and all results used the leave-one-out technique. As is shown in section 5, the results of the new classifier are comparable with the best other classifiers and the leave-one-out technique was not used. Lots of other research has used the datasets tested in section 5, for example [12], where they included some novel techniques for trying to remove noise or dirt [15] from the input dataset. The Wine dataset was tested again in [14], but with slightly worse results. For section 5.4, a new classifier was tried in [13] for classifying knowledge about web page use. While their classifier was superior, they made comparisons with a Bayes and a k-NN classifier that did not perform better than the new model suggested here.

One imaginary type of classifier that might be more similar could be a k-NN classifier that compares each dimension separately. So it would in fact be a number of 1-D comparisons and then a vote or count over them to determine the closest neighbours. This is not how the k-NN classifier is typically used however, but [3] tests a variation on this that uses subsets of features (dimensions) instead. That paper also notes the local and independent nature of the classifications in a k-NN. It also notes that training several classifiers using resampling or replication, for example, does not work as well with k-NN, which actually prefers a more discretely defined dataset. So while they do not suggest training different classifiers on modified dataset versions, they do suggest separating the variables into groups and creating different classifiers that way. Each variable group (1 to many) can produce a result that is summed with the other results to estimate an overall total.

## 3   The New Classifier Architecture

The decision to use a grid-like structure was to try to separate the input values, which would allow them to be more distinct in their relation to other sets of values. Adding another dimension here might help to simplify the learning process. It helps to visualise an artificial wave shape and therefore adds information that otherwise needs to be learned. Imagine a data row representing a number of variables, placed horizontally, and the bands for each variable then vertical below that. It also became clear that a typical set of real-valued inputs might not map easily to a 2-D or square topology. To map a single row of data values onto a 2-D grid therefore, requires the introduction of bands, where each value range is represented by a distinct cell in the band, or the second dimension. This is a bit like mapping an analogue value onto a set of discrete ones. If each value range is replaced by a graded cell, then the input can be placed approximately, but as it is distinct, it can store separate information. This deconstruction process would give it unique properties when linking up with data points from other variables. There is however, then a problem of how to map these cell sets onto the output categories. As each cell is more individual, it can be used to represent a more distinct relation with the output as well. This can be achieved by allowing each cell to store its own relation to the output categories and then combine or estimate over all of the relevant cells, when it comes to classifying something. The process uses a neural network-type of mechanism, by storing inherently in it, partial pieces of information that can be combined in a generic way. A traditional neural network (for example, [24]) stores the relation to the output by function transitions through layers of neurons. Only the output layer of those neurons is directly related to the output categories however, where the other layers are used to adjust the input, to make the final function transition the most suitable. For the new classifier of this paper, each band cell has one weight value to adjust the input and then a set of weights to represent its' importance to each output node. Each cell therefore has a single and direct relation to each output category, while in a neural network this is typically condensed into the output layer set of weights only. So it is these individual output weight sets that determine what the classifier result is, while the cell weight itself can still help with scaling. This is described again in section 5.1.

### 3.1    Classifier and Data Row Structure

The classifier and the data row representation share the same basic grid-like structure. A schematic of it is shown in Figure 1, where two rows of data points have been placed in their appropriate cells in the classifier. Each data row is therefore now a 2-D structure, where the continuous input values have been split up into discrete units. When training the classifier, each data row is added in turn and simply increments the weights in each cell that the values fall into. For each cell therefore, the scale weight and the related output category weight gets incremented. These combined value sets can then provide the generalisation properties of a neural network, for example. A major problem with existing classifiers is whether the data is linearly separable. That means – can you divide the categories using straight lines. If the data is not linearly separable, then more complex transformation functions are often required. This classifier might not have the same level of problem, as the linear construct is itself localised and deconstructed. Or possibly, the separating plane itself is broken-up by the localised nature of the cell to output values. So a strict differential curve might not be formed, but the separator could still be non-linear, which is something also written about in [5].

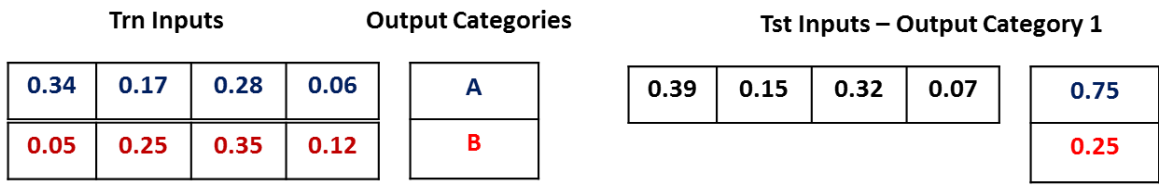

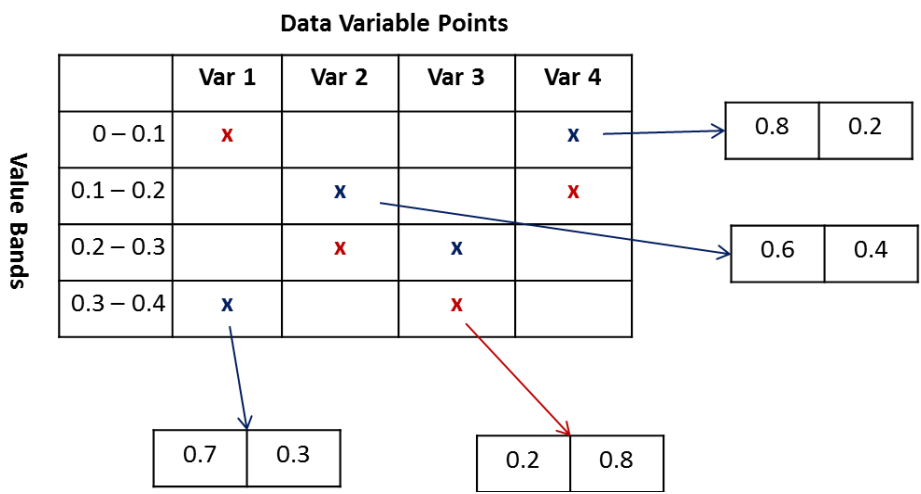

Figure 1. Classifier with 2 data rows displayed and output values

## 4   Example Scenario

Figure 1 is a schematic showing what a data row and the classifier itself might look like and a calculation from it. The top left of the figure shows two training dataset input rows, where there are only two categories to classify. Data row 1 belongs to category *A* and data row 2 belongs to category *B*. The data can be normalised first, which makes it easier to place into bands. For example, if the data is normalised to be in the range 0 to 1, then 10 bands of size 0.1 can easily be created. The bottom half of the schematic shows the classifier itself, where the two data rows are represented by using the same colour in each of the related cells. Each cell also stores its own relation to the output categories, which are made-up here, but show the expected weight bias. A new or test data row is presented, which falls into cells 1-4, 2-2, 3-4, 4-1. This is shown in the top right corner, with the supposed summed output weight values. For example, the first variable data point has the value of 0.39. This would fall into band 4 in the first grid column and therefore assign the value of 0.7 for category *A* and 0.3 for category *B*. The other 3 output sets are also illustrated. Looking at these sums shows that the new input row would be classified as belonging to category *A*.

The wave shape can be envisaged by joining up the red or the blue crosses in the data grid. This is for real values but binary data works just as well, because there is still a specific relation between each cell and the desired output category and the mapping is able to clearly define it.

## 4.1    Construction Process

For each training data row that is presented, the appropriate cell for each variable data point is calculated. Each selected band cell weight is then incremented, the output weight array for the cell is retrieved and the weight corresponding to the correct output category is also incremented. If the dataset is not well balanced with respect to number of rows in each category, then a bias can influence the resulting classification. So it is best if the training dataset has the same number of training data rows for each category type, but the weight increment amount for each output category can also be adjusted to try to remove the bias. See section 5.1 for more details on this. The training process requires only a single weight increment for each data row, without any subsequent adjustments and so any skewing of the relative increment amounts can have a significant effect.

## 5    Test Results

A test program has been written in the C# .Net language. It can read in a data file, normalise it, generate the classifier from it and measure how many row categories it subsequently evaluates correctly. A number of different tests have been carried out, to try to determine if the method can be flexible, or is able to generalise properly. It is still a shallow learning method at the moment, without the idea of a deep learning process through several levels of transformation. It is more visual in nature, providing an immediate feedback of what the dataset looks like. Three tests have been carried out. The first is simply to determine if the classifier works at all. The seconds tries to improve the results by fine-tuning some of the parameters. The third test considers other datasets that have a separate test dataset. Figure 2 shows an example weight set for a single variable '1' of the Iris [6] dataset. There were 12 bands, so each band would include a data point in the range of 0.08 or so, for example $0 - 0.08$, $0.081 - 0.16$, and so on. The input data points for the variable, after being scaled by the band weight, therefore produce the described values for each of the 3 output categories. It can be seen that the smaller bands (1 – 4) are mapped to category 1, some middle ones (5 and 6) are mapped to category 2 and the larger ones (7 - 12) to category 3.

| Variable 1: | | | Cat 1 | Cat 2 | Cat 3 |
|---|---|---|---|---|---|
| Weight (B1): | 0.06> | Outputs: | 0.18, | 0, | 0, |
| Weight (B2): | 0.047> | Outputs: | 0.14, | 0, | 0, |
| Weight (B3): | 0.167> | Outputs: | 0.4, | 0.08, | 0.02, |
| Weight (B4): | 0.12> | Outputs: | 0.22, | 0.14, | 0, |
| Weight (B5): | 0.14> | Outputs: | 0.06, | 0.26, | 0.1, |
| Weight (B6): | 0.1> | Outputs: | 0, | 0.2, | 0.1, |
| Weight (B7): | 0.087> | Outputs: | 0, | 0.1, | 0.16, |
| Weight (B8): | 0.147> | Outputs: | 0, | 0.16, | 0.28, |
| Weight (B9): | 0.053> | Outputs: | 0, | 0.06, | 0.1, |
| Weight (B10): | 0.033> | Outputs: | 0, | 0, | 0.1, |
| Weight (B11): | 0.013> | Outputs: | 0, | 0, | 0.04, |
| Weight (B12): | 0.033> | Outputs: | 0, | 0, | 0.1, |

Figure 2. Example set of weights for Variable bands.

## 5.1    Example Process and Parameter Values

The classifier might work well in a general sense, but there are still some configuration problems with it, especially for skewed or unbalanced data. One dataset that was used was the Wine [7] dataset. The dataset itself was slightly skewed, with 59 examples of category 1, 71 examples of category 2 and 48 examples of category 3. The cell increment value was typically set to *1 / number of rows* (a). As each row presentation increments the weight, this could be adjusted further to take account of the number of rows in each category. So it could be that category 1 would be incremented by *1/59*, category 2 by *1/71* and category 3 by *1/48* (b). Although, it was also the case that a weight increment for each output category of *1/50* worked just as well for the Wine dataset. Random update values however, for example, *1/71*, *1/48*, *1/59* respectively, would not work as well, but it was less sensitive if the scale value was the same across every category. During training therefore, a cell from the variable band is selected, based on the input data point size. That cell's weight is incremented by the value *(a)*. The correct output category is identified and the related weight value is incremented by the appropriate value *(b)*. When guessing the correct category, the input data point is placed in the appropriate cell and weighted by the cell weight, which is mainly for scaling. This weighted value is passed to the output array that weights it again, using the weight values for each output category. The other variables then update their selected bands in the same way. These output weights can then be summed for each category and the largest total value selected as the correct category. In Figure 2, for example, if the cell band was B5, the input value would firstly be weighted by 0.14, which could produce output suggestions of 0.06, 0.26 and 0.1 for the 3 categories. For that particular variable, category 2 would then be suggested. Basic algorithms for the process are described in Appendix A.

## 5.2    Test 1 – Initial Values

The classifier was tested on 3 classical datasets from the UCI Machine Learning Repository [23]. These were the Zoo database [25], Iris Plants database [6] and the Wine Recognition database [7]. These datasets have been used before as benchmark examples for testing new classifier designs. While they have been used to test self-organising classifiers, as they include output categories, they are suitable for testing the new supervised classifier as well. This first test was to determine if the classifier can correctly classify some data. The fixed parameters in this test give something to try to improve on in further tests. All of the datasets were assigned output category increment amounts as described in section 5.1, but it was really only the Zoo dataset that was too unbalanced. Strangely, the Wine dataset scored 177 correct from 178 without re-balancing, but the Zoo dataset really needed it. The Iris plant dataset assigns 50 rows to each category and so does not require any re-balancing in this respect. So in some cases, another decision can be made as to what the increment scaling should be, if there is a sampling problem, or whatever. If a poor re-balancing was applied however, this would make the values much worse and so the classifier is definitely sensitive to the increment scales. The following Table 1 gives the results of how many rows in each dataset the classifier correctly classified.

| Dataset Name | Correctly classified | % Correct |
|:---:|:---:|:---:|
| Zoo | 91 from 101 | 90% |
| Wine (scaling) | 172 from 178 | 96% |
| Wine (no scaling) | 177 from 178 | 99% |
| Iris | 143 from 150 | 95% |

Table 1. Classifier test results with 10 band cells for each variable.

For each dataset, the input data was firstly normalised to be in the range 0 to 1. Each data row was then presented to the classifier with the corresponding output category, and the appropriate cell values would be updated. After this training phase, each row would be presented again and the classifier would calculate its output category for that row; which would be compared to the correct category. So as all of the data rows have been combined into the same set of cells, it is not a direct retrieval of the input data, but still a generalisation over it. Note the two values for the Wine dataset, one with scaling the output increments and one without the scaling. Note also that each data row was presented only once and performed only 1 weight update on the appropriate cell.

## 5.3   Test 2 – Adjusting the Variable Parameters

Some other tests have been carried out to try to improve on the basic setup of the first test. The obvious change is to try to improve the band boundaries. The number of bands for each variable was both increased and decreased, and it was clear that an optimal number of bands could be determined, with a decrease in performance either side of that. There was also an attempt to try to learn better boundaries through additional classification processes. For example, separate the bands where the input data points have the largest gaps. So the band sizes themselves would be uneven and possibly different numbers of cells for each variable, but this did not give better results as a rule. Evenly spaced bands and the same number of cells for each input variable produced more reliable results. As changing the number however did affect the result, changing the grouping of data points is still important. This was coupled however with the weight increment value, which was even more critical and a better increment value would be more tolerant of the number of bands.

Other measurements considered the statistical features of each data column or variable separately. For example, with the Zoo [25] dataset with binary inputs, 2 bands actually provided as good results as a larger number. Therefore, the variance was considered as a possible indicator of the correct number of bands. In some cases this was useful and in others, less so. It was also found that there could be local optima when increasing the number of bands, where the classification percentage could drop and then increase again, as the band number increased. Or there could just be a single optimum. For one dataset, something like 100 bands would be much more accurate for the train dataset but less so for the test dataset. Therefore, it was possible to over-train by adding too many bands as well. Some new results are shown in Table 2, with the optimal number of band cells that were selected. The paper [12] tested the Iris, Wine and Zoo datasets using k-NN and neural network classifiers, with maybe 95.67%, 96% or 94.5% as the best results from one of the classifiers respectively. The values presented here are therefore at least equal to that.

| Dataset Name | Bands Number | Correctly classified | % Correct |
|:---:|:---:|:---:|:---:|
| Zoo | 2 | 94 from 101 | 93% |
| Wine (no scaling) | 15 | 178 from 178 | 100% |
| Iris | 12 | 145 from 150 | 96.7% |
| Abalone | 160 | 1452 from 4177 | 35% |

Table 2. Classifier Test results with optimal number of bands.

An example of a dataset that was not classified very well was the Abalone shellfish dataset [2]. That dataset had the most output categories at 29 and they were quite unevenly represented. It continued to improve slightly, with increasing numbers of cell bands, but in fact, the computer ran out of memory for even larger band numbers, than what is shown in Table 2. It was also helped a little bit by allowing the band boundaries to be learned. Their paper tried a decision tree C4.5, a k-NN nearest neighbour and a 1R classifier and reported 73% accuracy, compared to only 35% here. Another paper tested the dataset using k-means and a hierarchical clustering method. While k-means scored 62% accuracy, the hierarchical method scored only 6% accuracy and so the new classifier still looks quite adaptable.

## 5.4    Test 3 – Separate Train and Test Datasets

The User Modelling dataset [13] was used as part of a knowledge-modelling project that produced a new type of classifier. Their classifier was shown to be much better than the standard ones for the particular problem of web page use, classifying to 97.9% accuracy. This was compared to 85% accuracy for a k-NN classifier and 73.8 for a Bayes classifier. The training dataset however is quite skewed, with close to a factor of 4 between the least popular and the most popular output category. If using the frequency numbers exactly to re-balance, this still skewed the weight values, as the less popular output category can be assigned too large a weight update value. Therefore, the re-balancing itself was adjusted. In this case the weight update values were adjusted to something closer to: category 1 would be incremented by *1/34*, category 2 by *1/73*, category 3 by *1/78* and category 4 by *1/53*. This was an adjustment of +-10 on the actual frequency values. The new method of this paper does not classify as well as the 97.9% of their classifier, but it achieved 127 correct from 145, or 87.6% accuracy. This is with a separate test dataset to the train one and required 14 bands for each input variable. So here, the classifier works better than a Bayes classifier, for example.

Another test tried to classify the bank notes dataset [18]. These were scanned variable values from 'real' or 'fake' banknotes, where the output was therefore binary. This is another different type of problem, where a Wavelet transform might typically be used. The dataset again contained a train and a test dataset, where the best classification realised 81% accuracy, or 81 correct out of 100 in the test dataset. This was with 17 bands for each input variable. In that paper they quote maybe only 61% correct classification, but other papers have quoted close to 100% correct for similar problems.

## 6    Discussion and Future Work

As the classifier works well in a general sense, there must be differences with other classifiers. One difference is a set of weight values (bands) for each input variable. Each weight in a band then represents only a subset of the values that variables can take. So there is a changing weight value across the variable value range. Another difference is the separate sets of weights for each variable, making the classifier more orthogonal and even the batch treatment of the output categories. The band (or neuron) layer needs to be configured however and this is set from the start. Considering a multi-dimensional space that needs to be partitioned, a neural network comparison is reasonable. So while a traditional neural network is all about integrating the values into the same function, this classifier recognises and uses the distinct characteristics from the very start. The weight structure still allows those value sets to be used together. As the band number becomes very large, then memory can become a problem. It would be nice to represent this by a single (differential) function, but there are also separate sets of weights for the cell relation to the output categories and so a single function might not be possible. Some comparisons with a k-NN classifier can also be made, in particular, with the more discrete nature of k-NN.

If each set of neurons or bands then competes for the output category, could there be a level of self-organisation as well? It certainly introduces probabilities into the equation and an incorrect classification would require an update, maybe the cell to be changed. So each variable band could compete for the input value, while the selected output category is a function over all of them. After initial training that sets the initial weight values, the bands could therefore be allowed to change dynamically. So we would have a training phase, then an adjustment phase, before trying on a test dataset. For example, the classifier receives a previously unseen set of input points and places them in cells that represent an output category *A*, but the correct classification is category *B*. It therefore needs to move towards a category *B* output evaluation. The weight sets for all involved cells can be retrieved and a typical incremental weight move made, towards the correct category. For example, as in Figure 1, concerning the cells with arrows, if the correct category is known to be *B*, then there is only one variable with a positive value in that cell band and 3 variables with negative values in their respective bands for that output category. The procedure might therefore adjust the weights by reinforcing the single correct band (Var 3) and decrementing the other related 3 weight values (Var 1, Var 2 and Var 4). This would serve to increase the 0.8 output value for Var 3 and decrease the 0.6, 0.7 and 0.8 output values of the other variables. The process could be called competitive, if the variables that indicate the correct desired output are reinforced and the ones that indicate other outputs are decreased, for example.

## 7   Conclusions

This paper describes a new type of classifier that can be trained very quickly and is also very accurate. The function that is used in this version is linear; it is not obviously one of the known types of neural network – Associative or SOM, for example, but it incorporates many of their features. Its strength might lie in the direct but also deconstructed mapping between each cell and the output categories, without the need for complex transformations. Hidden layers are not removed completely however and are realised in a slightly different manner, as the separate band cell sets for each variable. This new classifier might also have an advantage when it comes to linearly separable or non-separable datasets, as that solution appears to be inherent in the structure. It appears to be critical that the weight update balance is correct and the number of bands appears to be more important than learning exact boundaries. Evenly spaced boundaries worked just as well. It is also worth noting that the results are largely based on the parameter settings and would be predictable, or the same, for every run that used the same settings. As it is predictable and as the earlier text has described, there are quite close similarities with a k-NN classifier as well. The test results show that the method can classify as well as or better than more established methods that include k-NN, Wavelet, Bayes, neural networks, discriminant analysis. It is unusual for a classifier to be better in a general sense and so it must be doing something new.

The model was realised through considering the wave shape design of [10], for example and it is more visual in nature than other designs. The extra dimension of the bands gives the structure a visual perspective of where the individual values might link together to form similar wave shapes. These could then be grouped together into common synapse links, or something, if considering a biological model, but that is another project. This was however the reason for the comparison with [10] and the classification itself may be more on appearance than some type of deeply learned function, such as the deep learning neural networks. So it is interesting that is can still classify quite accurately, but it may prefer data that is discrete over data whose values overlap, which would be in keeping with the design.

## Appendix A – Basic Algorithms

This appendix gives some very basic algorithms to describe the main processes of initializing, training and using the classifier.

### A. Initialise the Classifier and Parameters

The data and computer program needs to be initialised first, before the training algorithm can be run.

1. Pre-process the dataset to normalise the values into the range 0 to 1, for example.
2. Decide on the number of bands.
    a. For example, if 10 bands, then each band size is 0.1.
3. Decide on the cell weight increment ($cw$).
    a. For example, if there are 100 rows, make it 1.0 / 100.
    b. Each time a cell is used then, increment its weight by that amount.
4. Decide on the weight update ($ow$) for an output category ($oc$).
    a. For example, if a category is represented by 40 of the 100 rows, decide to increment the weight by 1.0 / 40 each time a row from that category is presented. Or if there are 2 categories, update any weight by 1.0 / 50.
    b. Each time a cell is used, update its related output weight $ow[oc]$ by that amount.
5. Create a cell to store its own weight ($cw$) and a set of output weights ($ow[oc]$).
6. Create a row of cells to represent the band values of a variable.
7. Create a list of rows, one for each variable, to represent the input dataset.
8. Initialise the values to 0, so that subsequent increments can be made.

**B.  Train the Network**

The classifier is trained essentially by incrementing the cell values that the inputs fall into.

1.  Present an input dataset row of values to the classifier.
2.  For each value in the row:
    a.  Retrieve the related variable list of cells.
    b.  Determine the band that the value falls into. For example, if the band width is 0.1 and the variable value is 0.45, then add it to cell 5 in the row.
    c.  Add weight increment (c*w*) to the cell's own weight.
    d.  Retrieve the output category *oc* the input belongs to and increment the related output weight by the appropriate value: cell output weight[i] += *ow[oc]*.

**C.  Classify a new Data Row**

The classifier is used by retrieving the cells that the input falls into, summing their weight values and using that to determine the most likely output category.

1.  Create an output value list, one value for each output category (*ov[oc]*) and initialise.
2.  Present the data row to the classifier.
3.  For each value in the row:
    a.  Retrieve the related variable list of cells.
    b.  Determine the band that the value falls into, as for training.
    c.  Retrieve the cell for that band position.
    d.  For each output weight in the cell, scale the output weight by the cell weight value.
    e.  For each output value category *ov[oc]*, add to it the related scaled output weight value times the input value.
    f.  Repeat this weighting and summation process for all of the variables and output categories.
4.  At the end, one of the output categories should have a larger value, so make that the selected category for the data row.